\documentclass[10pt,twocolumn,letterpaper]{article}

\usepackage{wacv}
\usepackage{times}
\usepackage{epsfig}
\usepackage{graphicx}
\usepackage{amsmath}
\usepackage{amssymb}

\wacvfinalcopy 

\ifwacvfinal\fi
\begin{document}

\title{SAD: Saliency-based Defenses Against Adversarial Examples}

\author{Richard Tran \hspace{1em}  David Patrick \hspace{1cm} Michael Geyer \hspace{1cm} Amanda S. Fernandez\\
Vision \& Artificial Intelligence Lab \\
Department of Computer Science \\
University of Texas at San Antonio 
}

\maketitle
\ifwacvfinal\thispagestyle{empty}\fi

\begin{abstract}
With the rise in popularity of machine and deep learning models, there is an increased focus on their vulnerability to malicious inputs.
These adversarial examples drift model predictions away from the original intent of the network and are a growing concern in practical security.
In order to combat these attacks, neural networks can leverage traditional image processing approaches or state-of-the-art defensive models to reduce perturbations in the data.
Defensive approaches that take a global approach to noise reduction are effective against adversarial attacks, however their lossy approach often distorts important data within the image.
In this work, we propose a visual saliency based approach to cleaning data affected by an adversarial attack.
Our model leverages the salient regions of an adversarial image in order to provide a targeted countermeasure while comparatively reducing loss within the cleaned images.
We measure the accuracy of our model by evaluating the effectiveness of state-of-the-art saliency methods prior to attack, under attack, and after application of cleaning methods.
We demonstrate the effectiveness of our proposed approach in comparison with related defenses and against established adversarial attack methods, across two saliency datasets.
Our targeted approach shows significant improvements in a range of standard statistical and distance saliency metrics, in comparison with both traditional and state-of-the-art approaches.
\end{abstract}

\section{Introduction} 

With increased adoption of machine and deep learning models into critical systems, adversarial attacks against these models have proportionally become a growing concern. 
Adversarial examples have been demonstrated in a growing range of applications, not only in classification tasks, but also in malware detection\cite{grosse2017adversarial} and recognition of speech and audio\cite{qin2019imperceptible}.
In the physical world, models used in facial recognition systems \cite{sharif2016accessorize} and autonomous vehicle interpretations of traffic signs, road patterns, and pedestrians \cite{kurakin2016adversarial} are also susceptible to these distorted inputs.
Adversarial attacks, such as fast gradient sign method (FGSM)\cite{FGSM}, iterative FGSM (I-FGSM)\cite{IFGSM}, Carlini-Wagner's L2 (CWL2)\cite{CWL2} and DeepFool\cite{deepfool}, take a broad range of approaches toward a similar goal: drifting a targeted model away its original intent through a series of malicious inputs\cite{athalye2018synthesizing}.
Attacks can be categorized as targeted or non-targeted, describing their intention of misleading a model toward a specific desired outcome or generally causing it to incorrectly interpret the input.
Additionally, attacks may be white- or black- box, if the adversary has prior knowledge of the underlying model, training information, or system.
\begin{figure}[hbt]
\centering
  \includegraphics[width=0.20\textwidth]{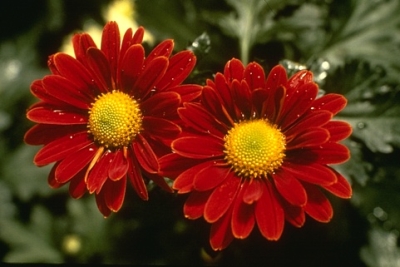}
  \includegraphics[width=0.20\textwidth]{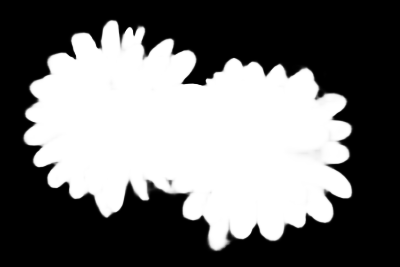}
  \includegraphics[width=1.7cm]{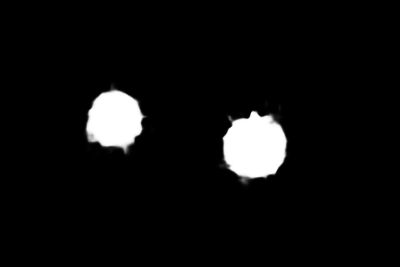}
  \includegraphics[width=1.7cm]{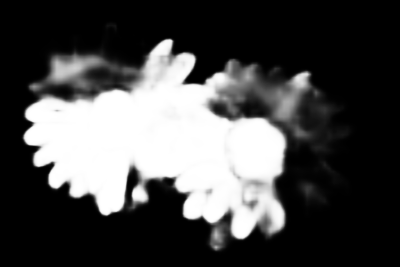}
  \includegraphics[width=1.7cm]{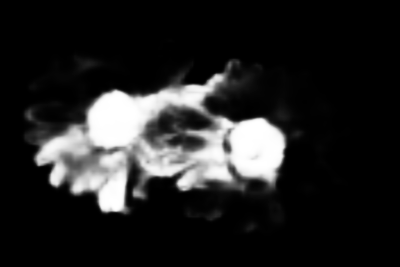}
  \includegraphics[width=1.7cm]{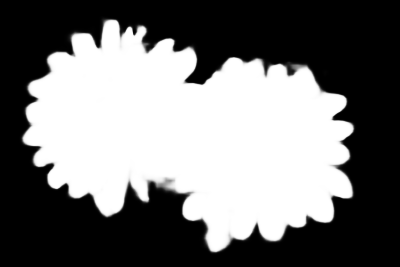}
  \caption{Comparison of CPD\cite{wu2019cascaded} on cleaning measures of ECSSD\cite{ECSSD} data attacked with FGSM\cite{FGSM}.
  Top row: Original image, ground truth saliency map;
  Bottom row from left (defenses): bit-depth reduction, JPEG compression, SHIELD\cite{das2018shield}, SAD.
  }
  \label{motivation}
\end{figure}
In this work, we focus on black-box non-targeted attacks to images, in an attempt to provide defense for general machine learning models against a range of attacks.
We leverage prevalent defense measures, identifying approaches that take a global approach to removing distortions as well as recent localized approaches.
Ultimately, we propose a new defense strategy, SAD, which uses regions of interest to strategically reduce adversarial distortions.
Figure \ref{motivation} motivates the effectiveness of our proposed approach, comparing the saliency maps of an adversarial example image after application of bit-depth reduction, JPEG compression, SHIELD\cite{das2018shield}, and our proposed SAD model.
Some defense techniques, including bit-depth reduction and JPEG compression, use a globalized approach to clean the inputs, while other techniques, such as SHIELD \cite{das2018shield}, use a more localized approach in reducing distortions.
Both types of approach have demonstrated performance against adversarial attacks of image classification models, however it is difficult to ensure preservation of the data integrity.

\begin{figure}[hbt]
\centering
  \includegraphics[width=0.20\textwidth]{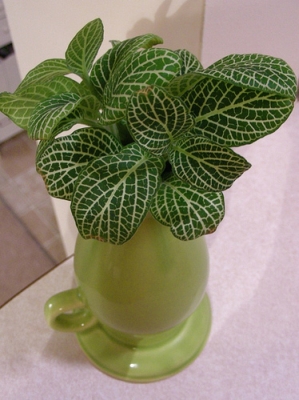}
  \includegraphics[width=0.20\textwidth]{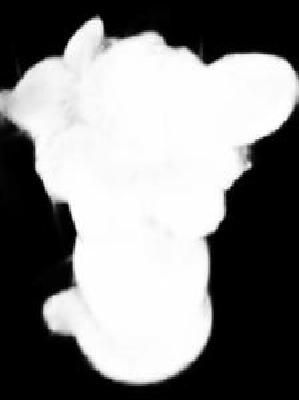}
  \includegraphics[width=0.20\textwidth]{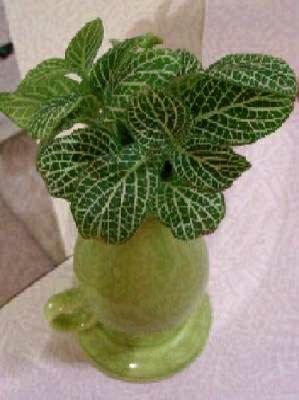}
  \includegraphics[width=0.20\textwidth]{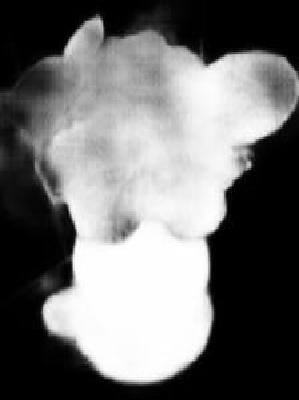}
  \caption{
  Consequences of adversarial attack on ECSSD data \cite{ECSSD} predicted using PiCANet \cite{liu2018picanet}. Top row: original image, Bottom row: FGSM attack \cite{FGSM}
  }
  \label{fig:sample_attack}
\end{figure}

While ROI region selection can be prohibitive on a perturbed image, some visual saliency models have been proven effective despite adversarial attack. \cite{adv_sal2019}.
In Figure \ref{fig:sample_attack}, an image from ECSSD\cite{ECSSD} dataset is shown with a generated saliency map, prior to and after a FGSM\cite{FGSM} attack.
Note the reduction in salient region identification in the upper leaves of the image, however the overall content remains correctly identified.



In this work, we propose a novel defense technique based on visual saliency.
The proposed approach identifies a region of interest (ROI) and leverages a saliency map to apply targeted cleaning techniques.
In demonstration of our proposed method, we evaluate the performance of state-of-the-art saliency models on established saliency data under the following conditions: original data, attacked by FGSM and DeepFool, and finally cleaned by four methods.
We discuss the impact of the choice of saliency estimation approaches in the effectiveness of our defense solution, and recommend augmentations for future improvement. 


\section{Related Work}
Among all the different cleaning techniques, we divide them into two major categories: globalized techniques and localized techniques. 
Globalized techniques, including methods such as bit-depth reduction and JPEG compression, have proven to be successful in reducing the effectiveness of adversarial attacks through the use of relatively simplistic approaches.
%
%
%
Bit-depth reduction limits the color in an image, which reduces distortions and therefore the effectiveness of the adversarial attacks.
However, while bit-depth reduction impacts general perturbations, it can also damage core features used to identify salient information.
%
JPEG Compression can also be used to reduce the effectiveness of malicious input by compressing the image.
This causes malicious inputs to get smoothed out, but at the same time introduces unwanted artifacts.
These unwanted artifacts can have unexpected consequences in saliency generation.
While each technique has a unique way of approaching the problem, they all reduce overall number of features within the data.
However, these globalized techniques are predictable and thus can be easily circumvented. 
%
%
Related to globalized approaches, distillation has also been demonstrated by Papernot et al as a viable means of defending against adversarial examples in deep neural networks\cite{Papernot2015DistillationAA}.
Magnet\cite{meng2017magnet} takes a cryptography-based approach to defending against adversarial examples.
Built for gray-box attacks, this defense is randomly selected from a set of precomputed methods at runtime.
Beyond globalized approaches, there are more localized approaches, such as image quilting \cite{quilting}, watermarking and SHIELD \cite{das2018shield}. Inherent randomness in these techniques makes them difficult for the adversarial attacks to circumvent.
%
%


We present a unique method that reduces the effectiveness of adversarial attacks while preserving original content, and demonstrate its viability by examining the saliency of the images prior to attacks, under attacks, and after defenses.



\section{Saliency-based Adversarial Defense (SAD)}
\begin{figure}[h]
  \includegraphics[scale=0.197]{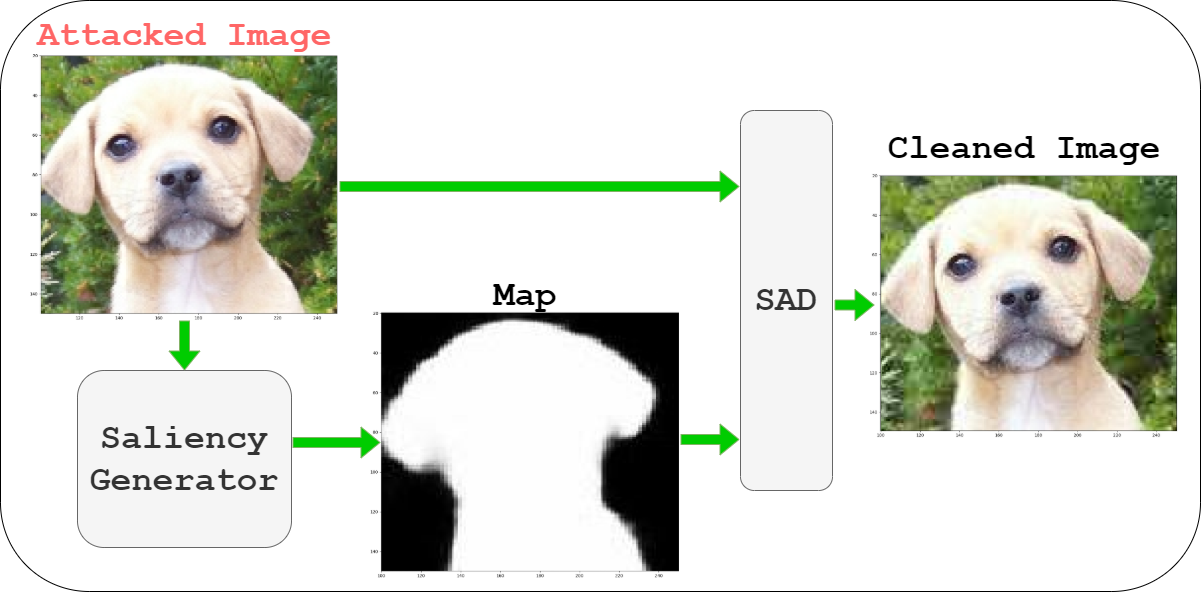}
  \caption{Overview of SAD
  }
  \label{fig:architecture}
\end{figure}

In response to the need for a targeted defense measure against diverse adversarial inputs, we propose a Saliency-based Adversarial Defense (SAD) approach outlined in Figure \ref{fig:architecture}.
Our model estimates relevant regions of interest (ROI) in an input and strategically applies countermeasures against adversarial perturbations.

\subsection{Model Description}
In order to select the most relevant ROI, our proposed model leverages a model for visual saliency estimation.
First, a saliency map is generated for the input image. Our implementation uses PiCANet \cite{liu2018picanet} trained on the DUTS-TR dataset \cite{Wang_2017_CVPR}. Once a map is generated, JPEG compression is applied at differing qualities based on the saliency predictions.

In addition to the images to be processed, a list of compression level must be passed as a parameter to our model. This list will be denoted as $Q$, where $Q(i)$ denotes the $i$th compression level.
Much like SHIELD \cite{das2018shield}, each image processed is segmented into 8$\times$8 windows. $W_{ij}$ is used to denote the window at the $i$th row and $j$th column of the image.
The saliency map, taken as a grey-scale image, is identically segmented into 8$\times$8 windows. Each window in the saliency map is assigned a scalar value, from 0 to 255, based on the average saliency prediction of all pixels within the window. These scalar values, denoted as $Sal_{ij}$, are then divided by a threshold and used as an index into $Q$. The compression level $C$ for $W_{ij}$ can be expressed as the following equation.

\begin{equation}
    C(W_{ij}) = Q( \left \lfloor{ \frac{Sal_{ij}\cdot|Q|}{255}}\right \rfloor  )
\end{equation}

The goal of this approach is to reduce the effectiveness of an adversarial attack while minimizing damage done to the ROI.

\begin{figure}[hbt]
\centering
  \includegraphics[scale=0.25]{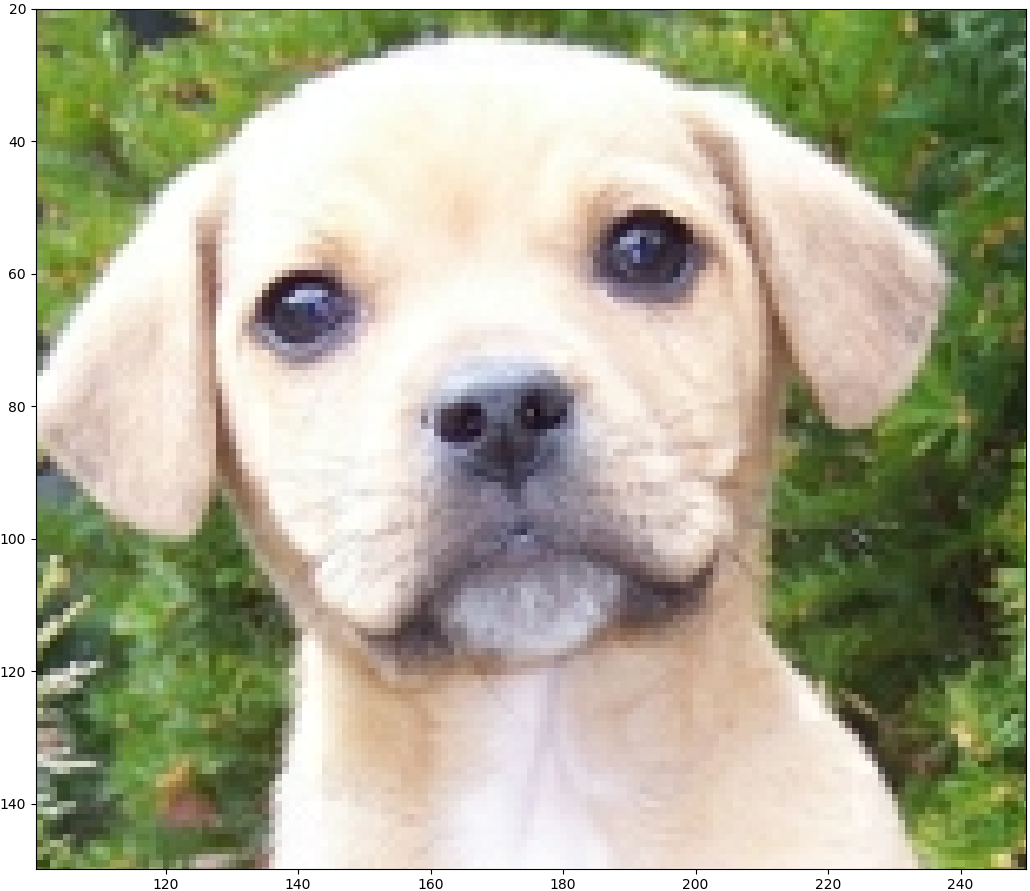}
  \includegraphics[scale=0.25]{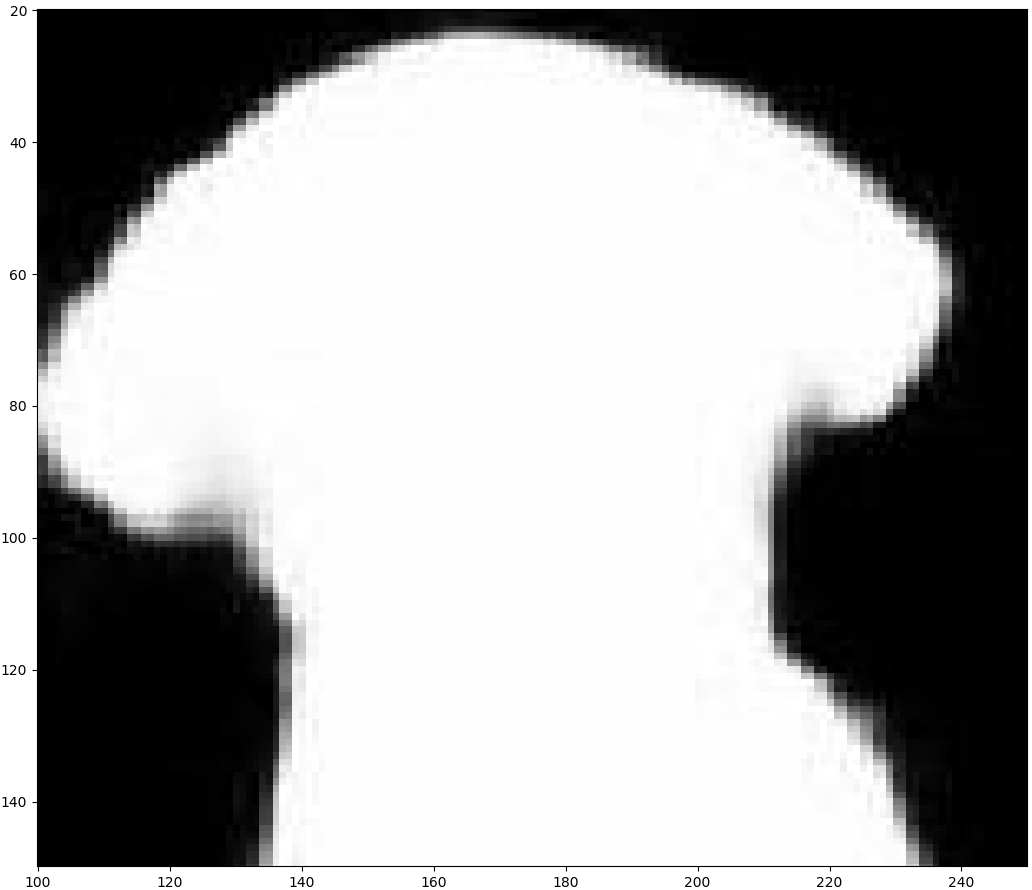}
  \includegraphics[scale=0.25]{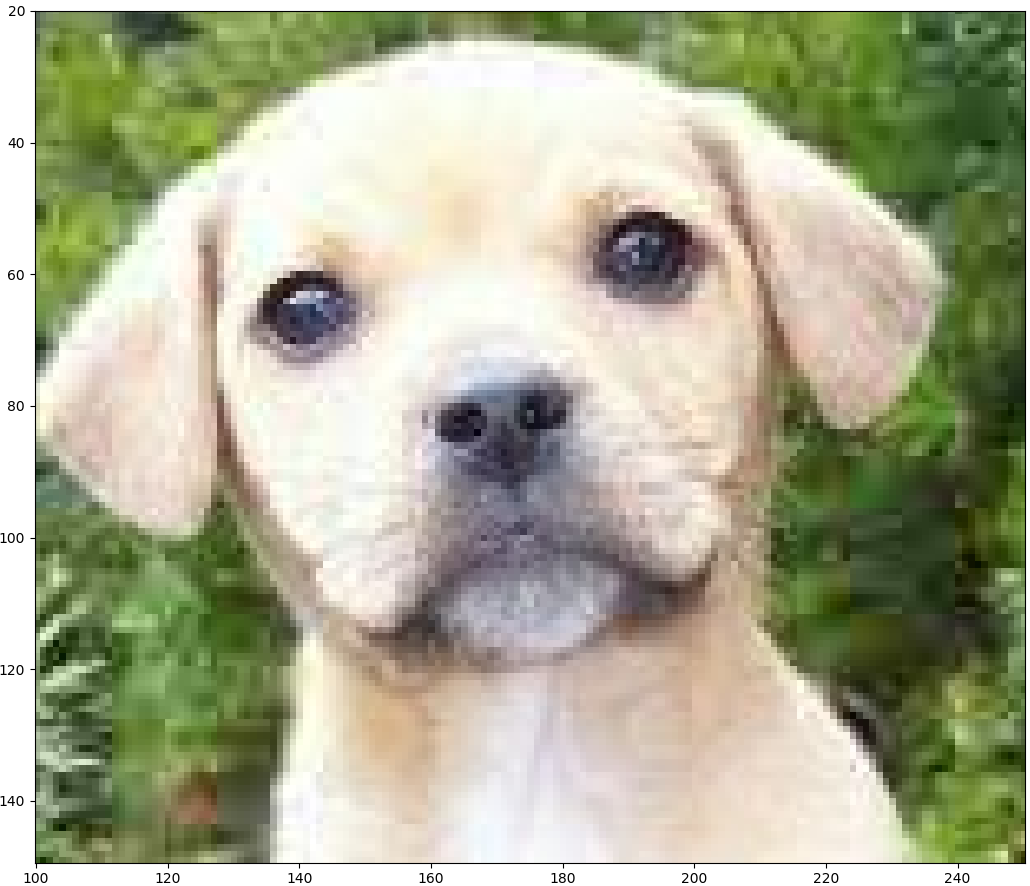}

  \caption{
  Example input and output of SAD. Top down: Original image, Saliency prediction, Output.
  }
  \label{fig:sample_defense}
\end{figure}

This technique is similar in nature to SHIELD \cite{das2018shield}, with the primary difference being the replacement of a randomized compression algorithm with a saliency based compression algorithm.
Figure \ref{fig:sample_defense} provides an example input, saliency prediction and output that demonstrates compressing the background significantly more than the salient parts of the image. In this example, the output has salient regions compressed at 90\% while non-salient regions are compressed at 20\%.


\section{Experiments \& Results}
In this section, we first review the extensive setup of datasets, adversarial image generations, adversarial defenses, and evaluation metrics used in this work.
We then specify the series of experiments performed, and demonstrate the performance of our proposed approach in the final section.

\subsection{Setup}

Two popular saliency datasets, ECSSD \cite{ECSSD} and SALICON \cite{jiang2015salicon}, were chosen for experimental setup, due to their prominence recent saliency research and their degree of difficulty. 
The Extended Complex Scene Saliency Dataset (ECSSD) \cite{ECSSD} is comprised of complex scenes, presenting textures and structures common to real-world images.
ECSSD \cite{ECSSD} contains 1,000 intricate images and respective ground-truth saliency maps, created as an average of the labeling of five human participants.
The Salicency in Context (SALICON) \cite{jiang2015salicon} is a similarly complex dataset, chosen in this work to provide a broader range of scenes and a larger number of samples.
SALICON \cite{jiang2015salicon} contains a training set of 10,000 images and their respective ground-truth saliency maps as well as a validation set of 5,000 images with their corresponding ground-truths and was designed to purpose of evaluating current saliency models with natural scene images.
In addition to ground truth saliency maps, this dataset provides fixation maps for analysis. 
For our experiments, we selected only the training set of SALICON \cite{jiang2015salicon} to evaluate the saliency models, for a total of 10,000 images on this dataset.

Two adversarial attacks were chosen in this work, for their prominence as well as diversity in approach.
We leveraged FGSM \cite{FGSM} and DeepFool \cite{deepfool} attacks to evaluate the efficacy of our proposed countermeasure.
The Fast Gradient Sign Method (FGSM) \cite{FGSM} was chosen as it is a more traditional attack which has proven to be effective in creating input images which are significantly misleading to popular convolutional frameworks. 
The inclusion of FGSM \cite{FGSM} allows us to test the effectiveness of saliency models and cleaning algorithms against a well-known and common adversarial attack. 
DeepFool \cite{deepfool} was chosen as it is considered a state-of-the-art attack against image-based classification models, with a more robust attack surface.
As each of these approaches requires an objective function to consider in their attack, we chose a common VGG16\cite{vgg16} backbone and pretrained this model using the ImageNet \cite{imagenet_cvpr09} dataset.

In defense against the adversarial examples, we selected three countermeasures in comparison with our proposed SAD approach: bit-depth Reduction, JPEG-compression, and SHIELD. \cite{das2018shield}.
We chose these as a balance of global and localized defense techniques, for a robust comparison with our proposed approach.
Both bit-depth reduction and JPEG-compression are established countermeasures to defend against adversarial attacks, effective in reducing the number of perturbations present within the images. 
For the purposes of our experiments, we used a 3-bit depth reduction and a compression level of 80 for JPEG-compression.
The recent Secure Heterogeneous Image Ensemble with Localized Denoising (SHIELD) \cite{das2018shield} uses a randomized compression levels to reduce the number of perturbations present within the images.

In order to evaluate the effectiveness of our defense, we put all images - original, adversarial, and "cleaned" - through state-of-the-art saliency models, and evaluate the performance of each model.
As these models have demonstrated top performance on these popular saliency datasets, we can establish how they are affected by the adversarial inputs.
In this work, we selected three diverse models to generate saliency maps for the images: BASNet \cite{BASNet}, CPD \cite{wu2019cascaded}, and SalGAN \cite{Pan_2017_SalGAN}. 
The Boundary-Aware Salient Object Detection model (BASNet) \cite{BASNet} uniquely leverages edges and bounding boxes to help establish a saliency map for an image.
The Cascaded Partial Decoder (CPD) \cite{wu2019cascaded} model incorporates a holistic attention mechanism into the traditional encoder-decoder framework. 
The Saliency GAN (SalGAN) \cite{Pan_2017_SalGAN} model is a generative adversarial network approach, providing discriminator and generator models in adversarial training.
It is important to note that SalGAN \cite{Pan_2017_SalGAN} was mainly designed with to generate saliency maps based on eye-fixations rather than a basic saliency map. 

Finally, we leverage saliency metrics in this work as an evaluation of the effectiveness of our proposed defense.
The MIT Saliency Benchmark \cite{mit-saliency-benchmark} provides established metrics for saliency estimation models, on both binary saliency maps and fixation maps.
For the purposes of this work, in application to only ground truth maps, we selected Earth Mover's Distance (EMD), Pearson's Correlation Coefficient (CC), Normalized Scanpath Saliency (NSS), KL-Divergence (KLD) and similarity score (SIM).


\subsection{Experiments}
To establish a baseline, we generated the saliency maps for the ECSSD and SALICON datasets using the BASNet\cite{BASNet}, CPD\cite{wu2019cascaded}, and SalGAN\cite{Pan_2017_SalGAN} saliency models.

After establishing a baseline, we then took each dataset and performed separate FGSM and DeepFool attacks on the images.
In uniform comparison, all attacks leveraged a VGG-16 common backbone.
The same saliency models were used to generate saliency maps of each set of these attacked images.

Finally, all attacked images were cleaned using a series of adversarial attack defenses.
We started by performing a bit-depth reduction on the attacked images, reducing the images to a 3-bit color representation.
Next, we performed a JPEG compression on the attacked images, compressing the image quality by 80 percent. 
Once we finished the JPEG compression, we performed the state-of-the-art defense SHIELD on the attacked images. SHIELD functions takes JPEG compression but instead applying a uniform image compression level, the compression is applied in patches, randomly determining the quality reduction of the image.  
We took the new images from all the current cleaning techniques and then fed them into all of the saliency models to get their respective saliency maps to use for the metrics.
Once we had all of the saliency maps, we then ran all of our metrics on the cleaned images to show how the cleaning techniques affected the saliency maps. 

Finally, using the same experimental guidelines, we performed SAD on the attacked datasets. 
Testing was performed with 2 lists of compression qualities (20, 50, 70, 70, 80, 90) and (50, 70, 90).
The cleaned SAD images were then run through the same metrics in order to make a direct comparison of our technique and other modern cleaning techniques.

\subsection{Results} 
\begin{figure}[hbt]

  \includegraphics[width=2cm, height=2cm]{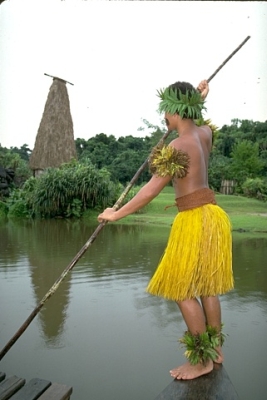}
  \includegraphics[width=2cm, height=2cm]{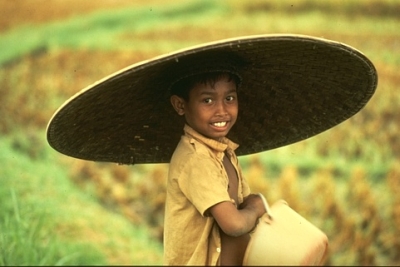}
  \includegraphics[width=2cm, height=2cm]{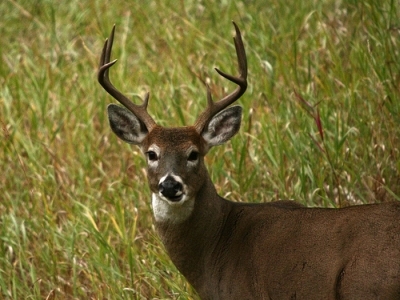}
  \includegraphics[width=2cm, height=2cm]{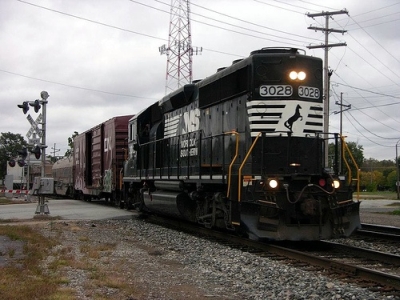}

  \includegraphics[width=2cm, height=2cm]{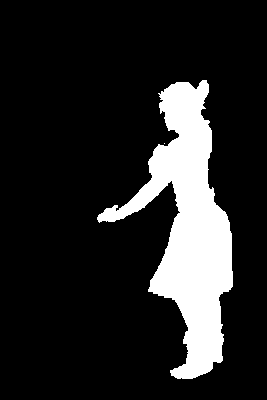}
  \includegraphics[width=2cm, height=2cm]{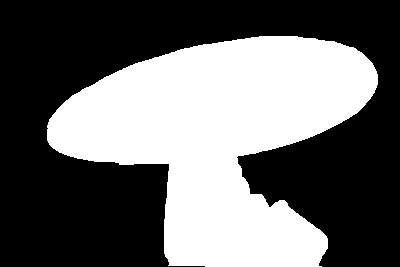}
  \includegraphics[width=2cm, height=2cm]{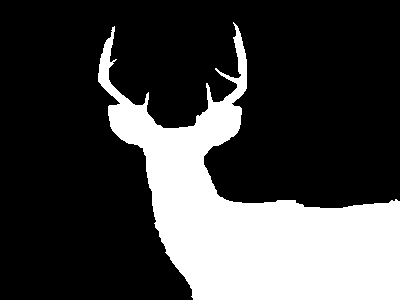}
  \includegraphics[width=2cm, height=2cm]{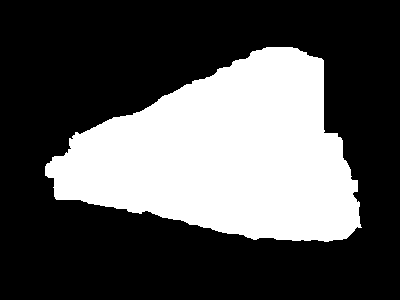}
  
  \includegraphics[width=2cm, height=2cm]{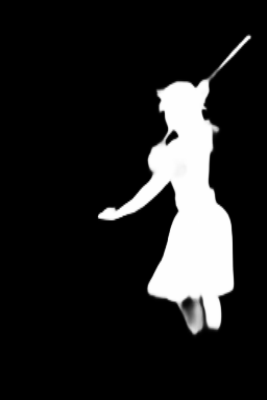}
  \includegraphics[width=2cm, height=2cm]{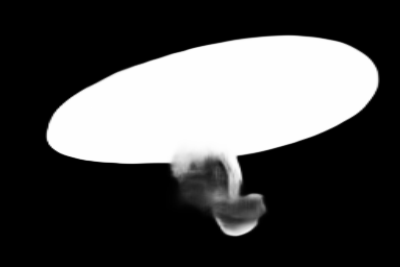}
  \includegraphics[width=2cm, height=2cm]{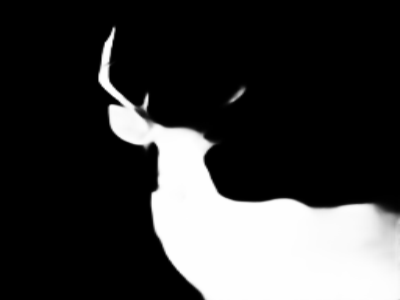}
  \includegraphics[width=2cm, height=2cm]{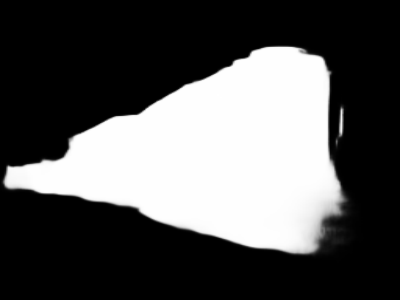}
  
  \includegraphics[width=2cm, height=2cm]{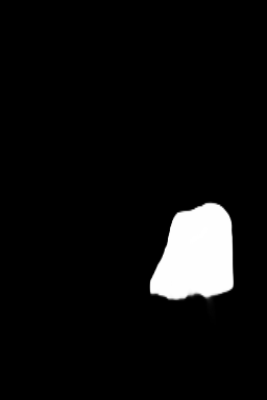}
  \includegraphics[width=2cm, height=2cm]{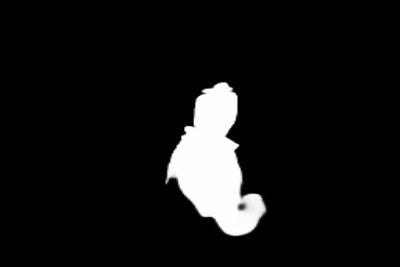}
  \includegraphics[width=2cm, height=2cm]{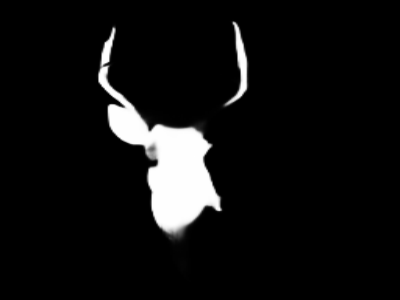}
  \includegraphics[width=2cm, height=2cm]{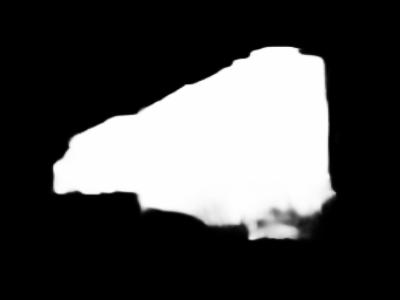}
  
  \includegraphics[width=2cm, height=2cm]{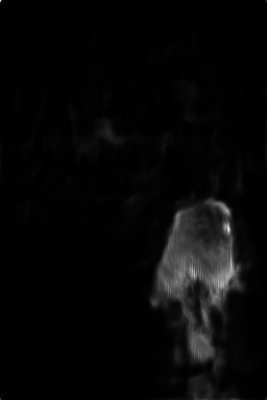}
  \includegraphics[width=2cm, height=2cm]{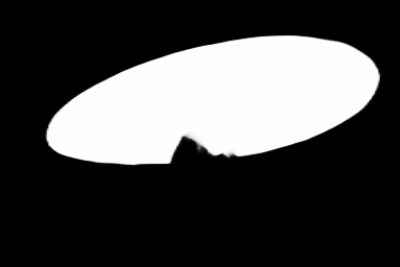}
  \includegraphics[width=2cm, height=2cm]{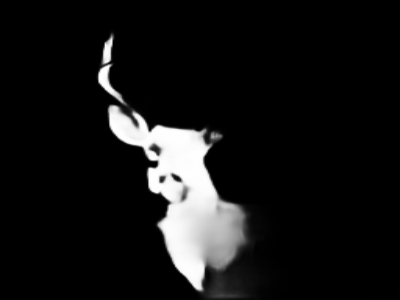}
  \includegraphics[width=2cm, height=2cm]{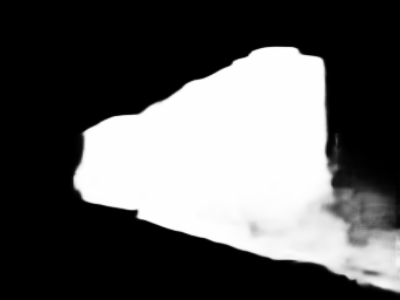}
  
  \includegraphics[width=2cm, height=2cm]{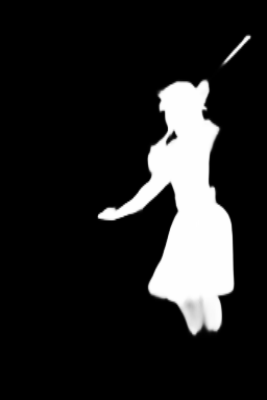}
  \includegraphics[width=2cm, height=2cm]{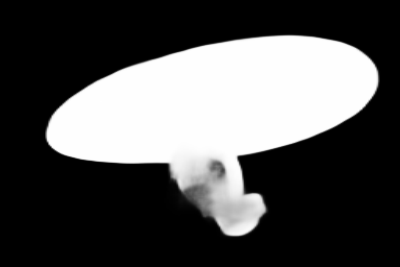}
  \includegraphics[width=2cm, height=2cm]{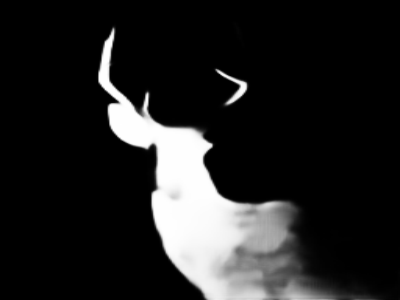}
  \includegraphics[width=2cm, height=2cm]{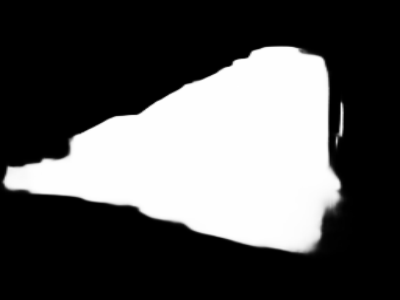}
  
    \includegraphics[width=2cm, height=2cm]{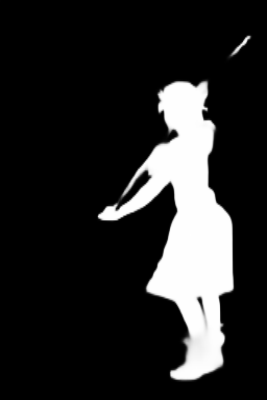}
    \includegraphics[width=2cm, height=2cm]{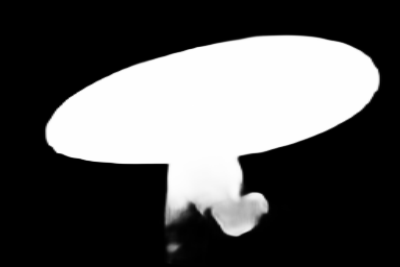}
    \includegraphics[width=2cm, height=2cm]{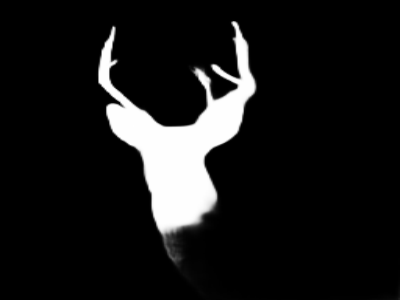}
    \includegraphics[width=2cm, height=2cm]{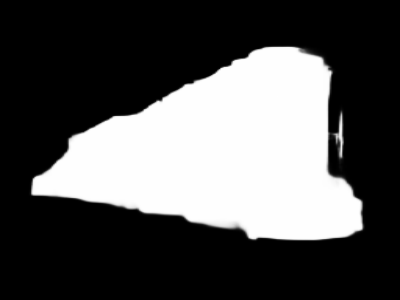}

  \caption{Comparison of BASNet\cite{BASNet} on cleaning measures of ECSSD\cite{ECSSD} data attacked with FGSM.
             From top: Original image, ground truth, Attacked, BitDepth, JPEG, SHIELD\cite{das2018shield}, SAD
  }
  \label{fig:results_ecssd}
\end{figure}

Figure \ref{fig:results_ecssd} provides a collection of images picked from ECSSD\cite{ECSSD}. The first two rows of the figure contain the original image and its respective ground truth. The third row contains the saliency map generated by BASNet\cite{BASNet} from an FGSM\cite{FGSM} adversarial example. In this figure, FGSM\cite{FGSM} is shown to cause minor distortions to the saliency maps that were generated by BASNET\cite{BASNet}. The following rows contain the resulting saliency map from the bit-depth reduction, JPEG-Compression, SHIELD\cite{das2018shield} and SAD respectively. These rows demonstrate the highlight the effects that each cleaning technique has on the saliency map generation. 

Table \ref{table:results_salicon_basnet} shows the results of running the BASNet\cite{BASNet} visual saliency model \cite{BASNet} against the SALICON\cite{jiang2015salicon} dataset. 

\begin{figure*}[hbt]
    \centering
    \includegraphics[scale=0.45]{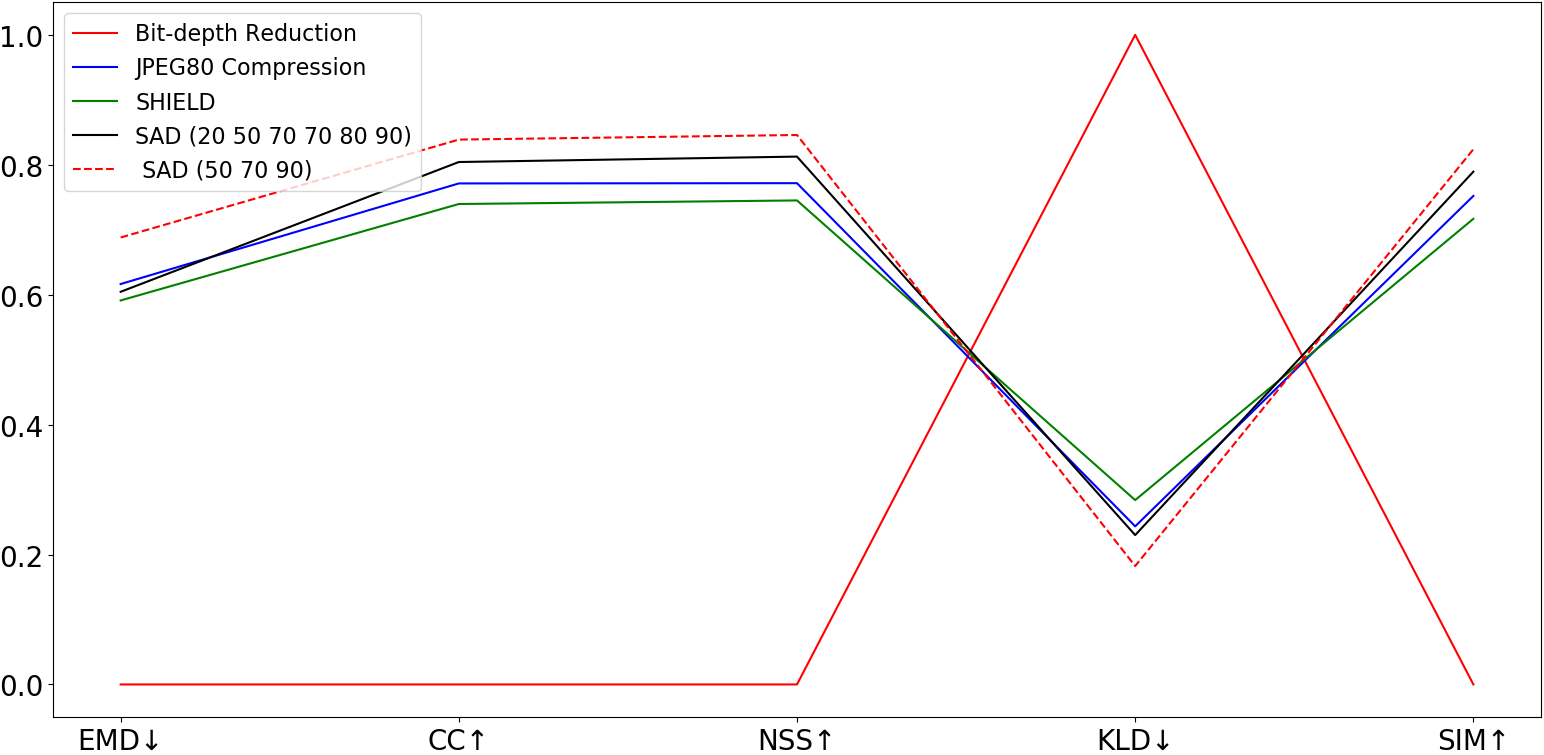}
    \caption{ECSSD \cite{ECSSD} attacked with FGSM \cite{FGSM} generated by BASNet \cite{BASNet} (min-max normalized) }
    \label{fig:ecssd_fgsm_basnet_metric_graph}
\end{figure*}

\begin{figure*}[hbt]
    \centering
    \includegraphics[scale=0.45]{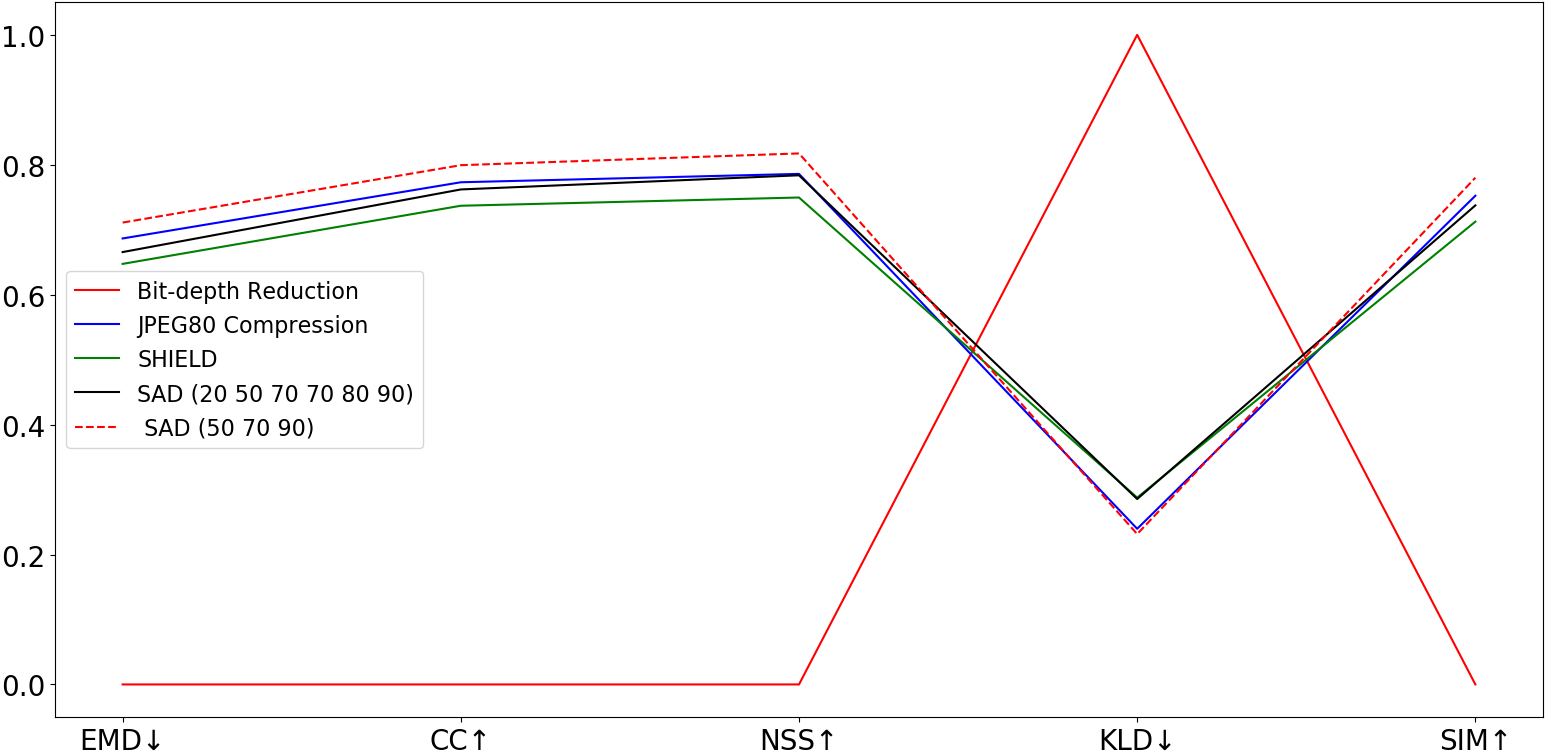}
    \caption{ECSSD \cite{ECSSD} attacked with FGSM \cite{FGSM} generated by CPD \cite{wu2019cascaded} (min-max normalized)}
    \label{fig:ecssd_fgsm_cpd_metric_graph}
\end{figure*}

Tables \ref{table:results_ecssd_basnet} and \ref{table:results_ecssd_cpd} show metric results of running BASNet\cite{BASNet} and CPD\cite{wu2019cascaded} respectively on the entire ECSSD\cite{ECSSD} dataset. In each of these cases we conclude that SAD performs significantly better on global attacks, such as FGSM\cite{FGSM}, than localised attacks, such as DeepFool\cite{deepfool}. This is because more distortions are present in the non-salient regions of global attacks, thus more distortions are removed overall. We posit that for localised attacks on ECSSD\cite{ECSSD}, while SAD performed worse than standard JPEG compression, the difference in performance is comparatively small. Figures \ref{fig:ecssd_fgsm_basnet_metric_graph} and \ref{fig:ecssd_fgsm_cpd_metric_graph} are min-max normalised graphs presented to visualise these results.

Table \ref{table:results_salicon_cpd} shows metric results of running CPD\cite{wu2019cascaded} on the SALICON\cite{jiang2015salicon} dataset. In this case, because CPD\cite{wu2019cascaded} does not perform well on this fixation based dataset, the overall results do not vary much between the original, attacked, and cleaned examples.

\begin{figure}[hbt]

  \includegraphics[width=2cm, height=2cm]{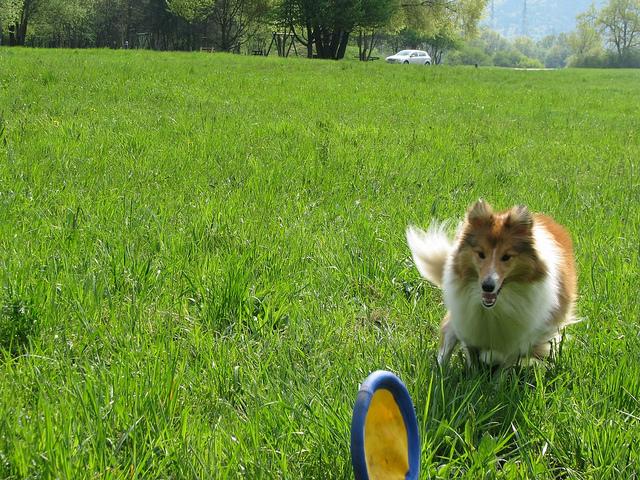}
  \includegraphics[width=2cm, height=2cm]{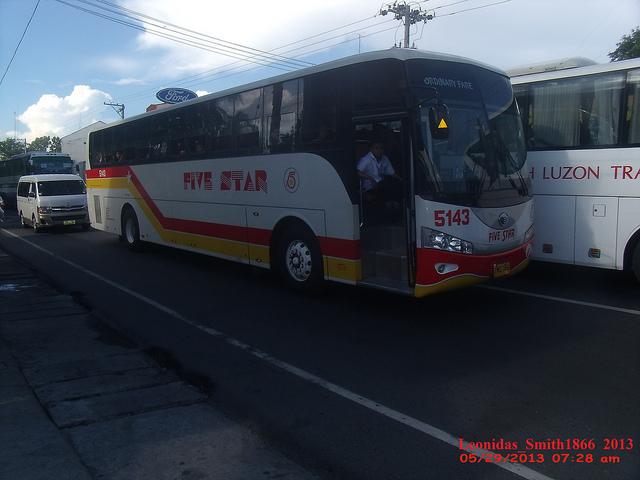}
  \includegraphics[width=2cm, height=2cm]{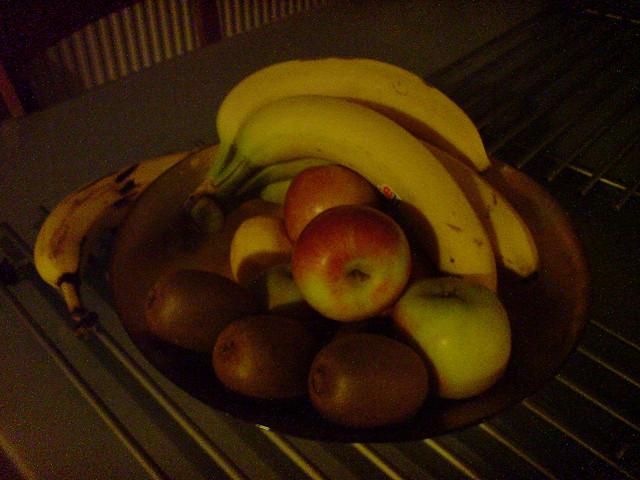}
  \includegraphics[width=2cm, height=2cm]{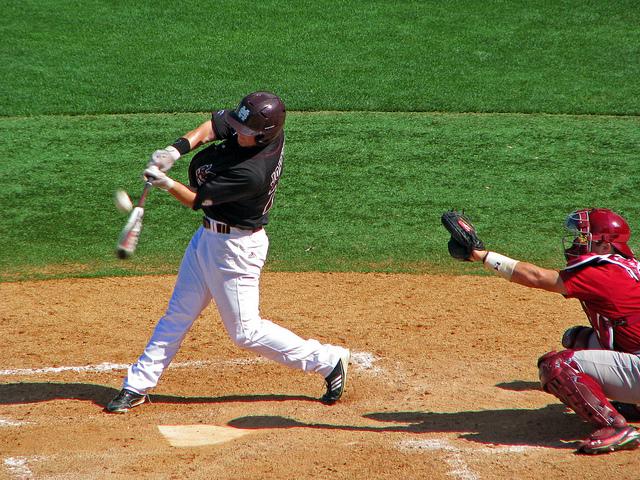}
  
  \includegraphics[width=2cm, height=2cm]{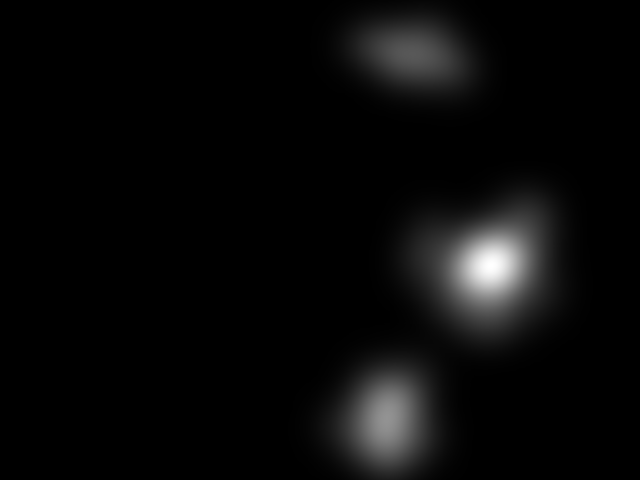}
  \includegraphics[width=2cm, height=2cm]{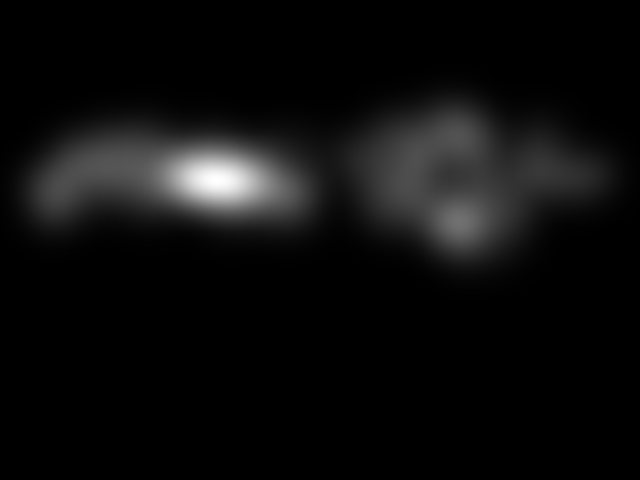}
  \includegraphics[width=2cm, height=2cm]{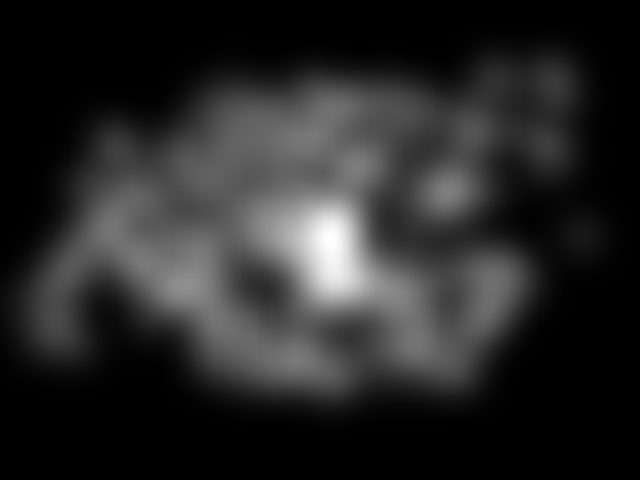}
  \includegraphics[width=2cm, height=2cm]{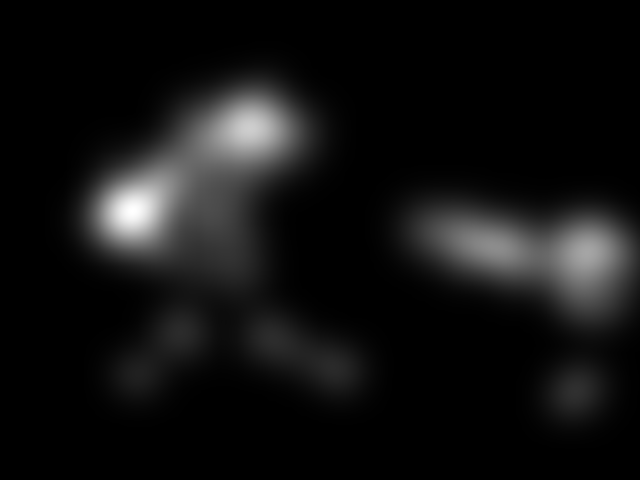}
  
  \includegraphics[width=2cm, height=2cm]{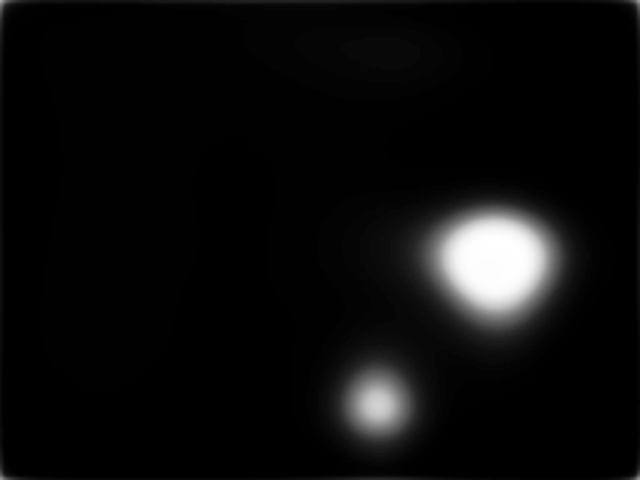}
  \includegraphics[width=2cm, height=2cm]{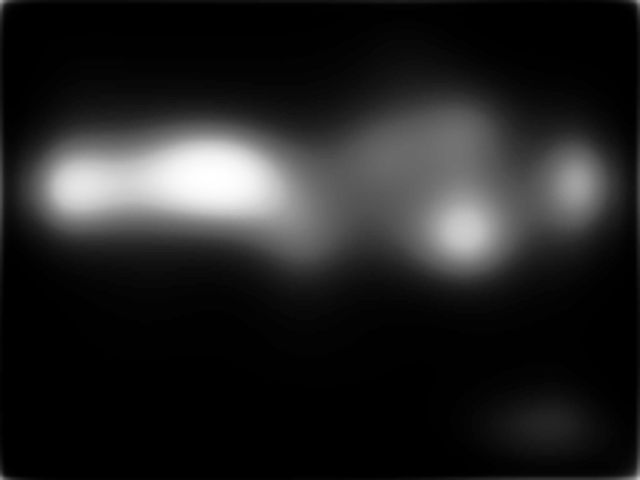}
  \includegraphics[width=2cm, height=2cm]{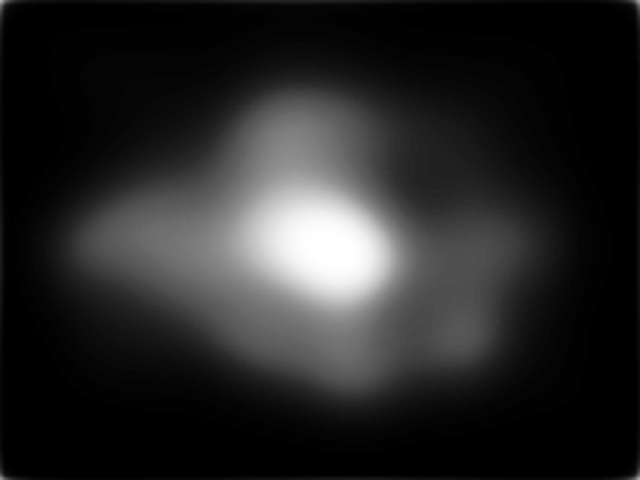}
  \includegraphics[width=2cm, height=2cm]{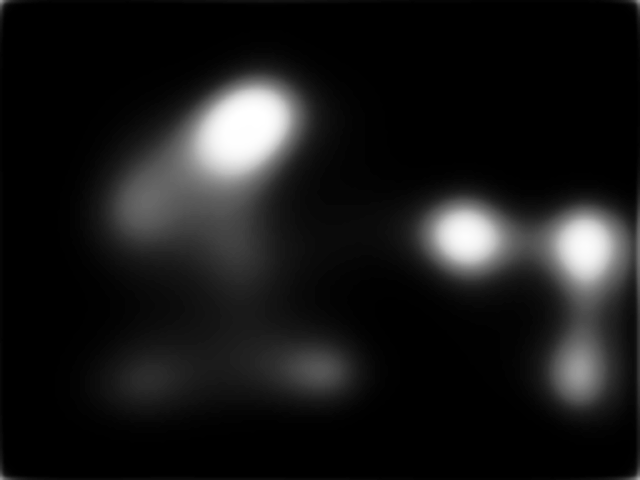}
  
  \includegraphics[width=2cm, height=2cm]{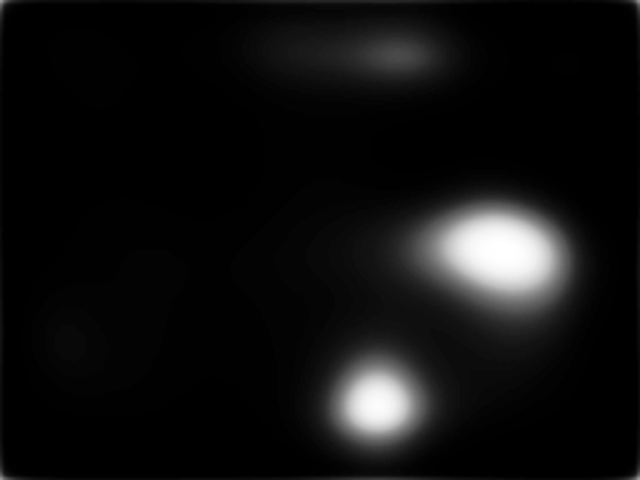}
  \includegraphics[width=2cm, height=2cm]{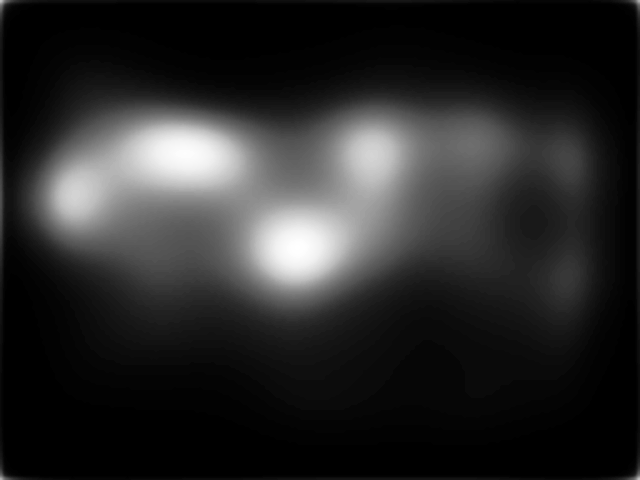}
  \includegraphics[width=2cm, height=2cm]{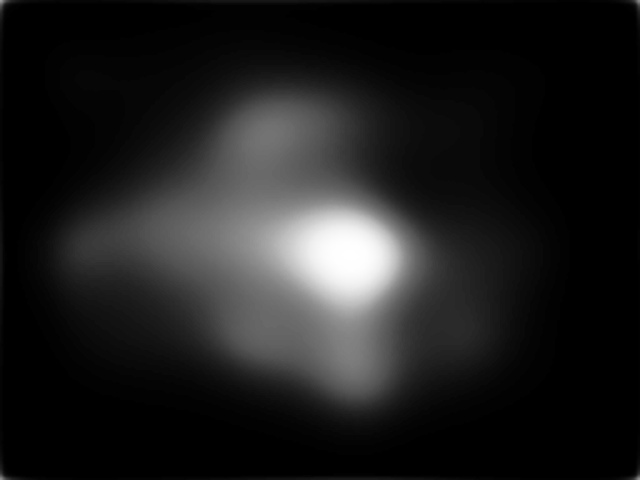}
  \includegraphics[width=2cm, height=2cm]{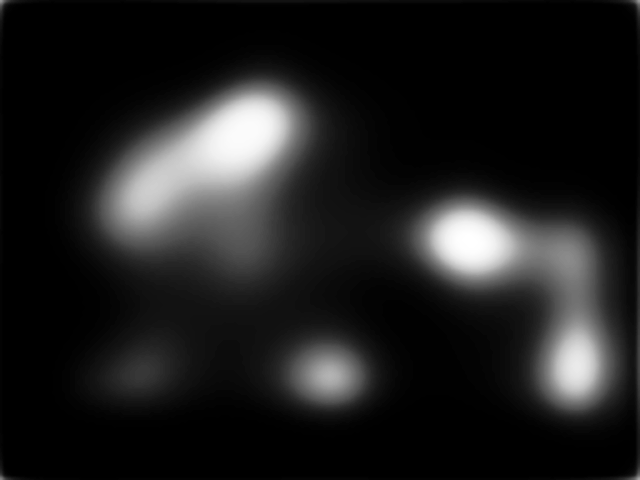}
  
  \includegraphics[width=2cm, height=2cm]{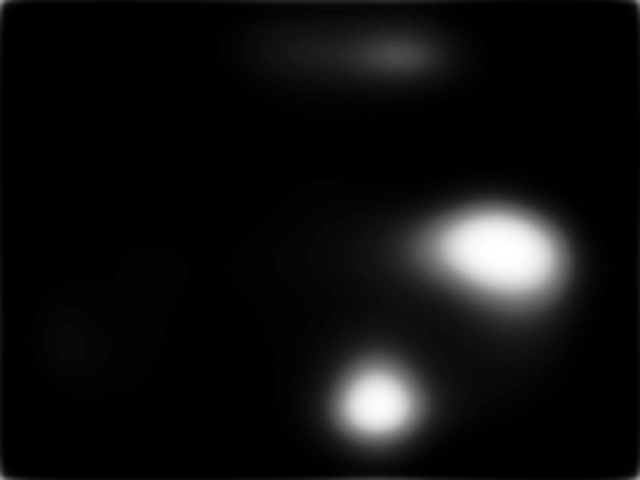}
  \includegraphics[width=2cm, height=2cm]{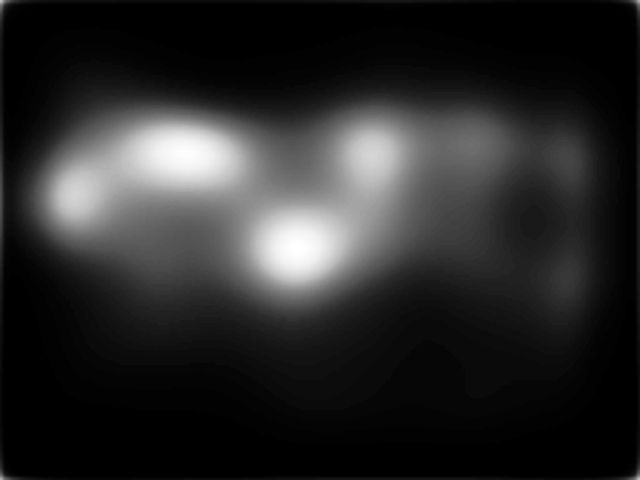}
  \includegraphics[width=2cm, height=2cm]{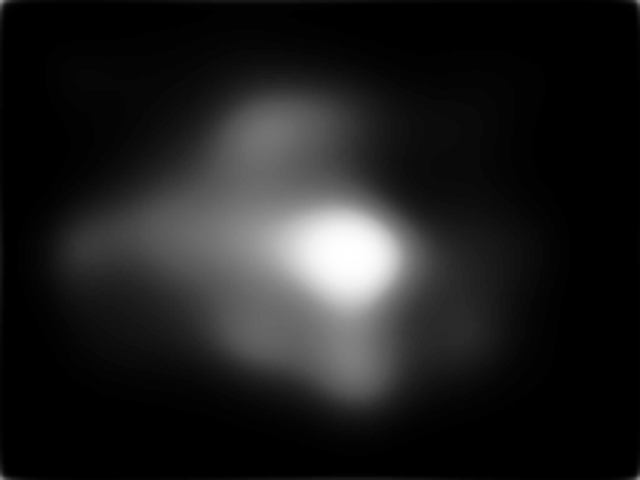}
  \includegraphics[width=2cm, height=2cm]{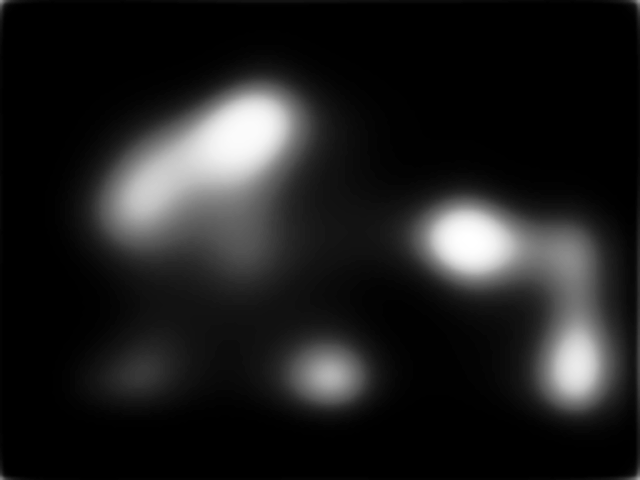}
  
  \includegraphics[width=2cm, height=2cm]{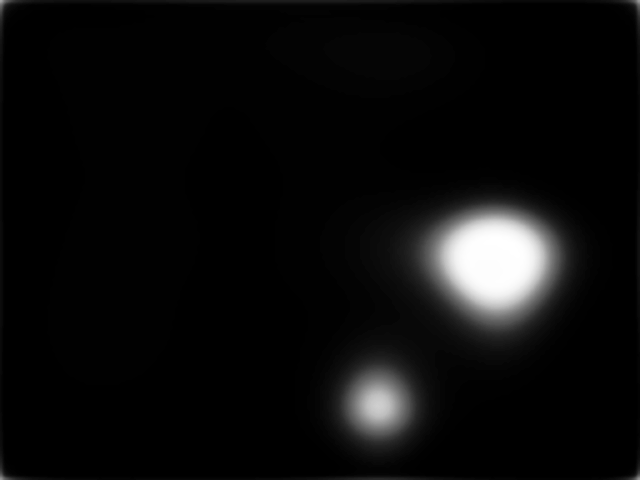}
  \includegraphics[width=2cm, height=2cm]{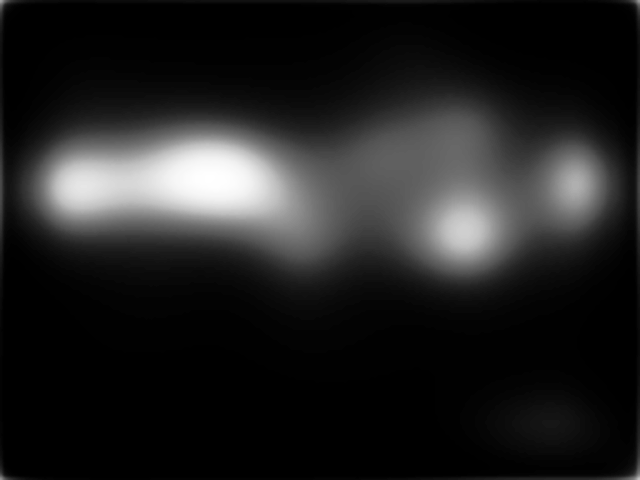}
  \includegraphics[width=2cm, height=2cm]{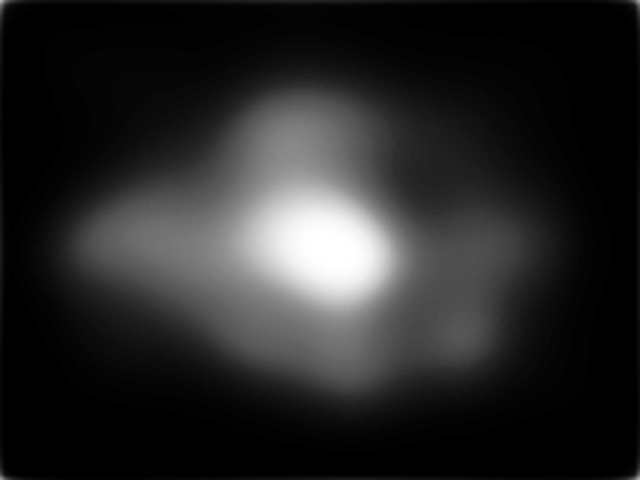}
  \includegraphics[width=2cm, height=2cm]{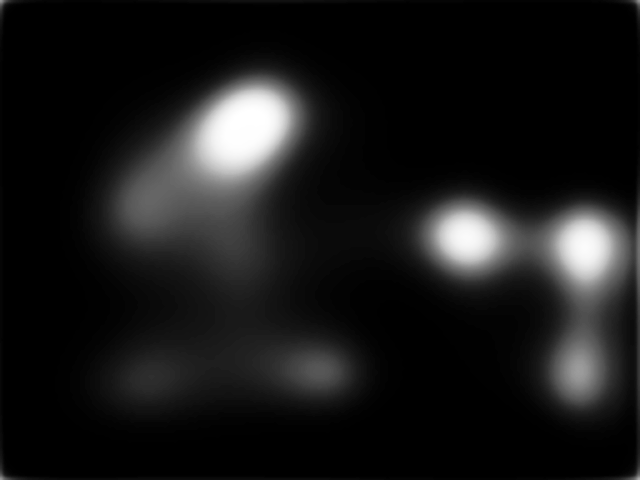}
  
  \includegraphics[width=2cm, height=2cm]{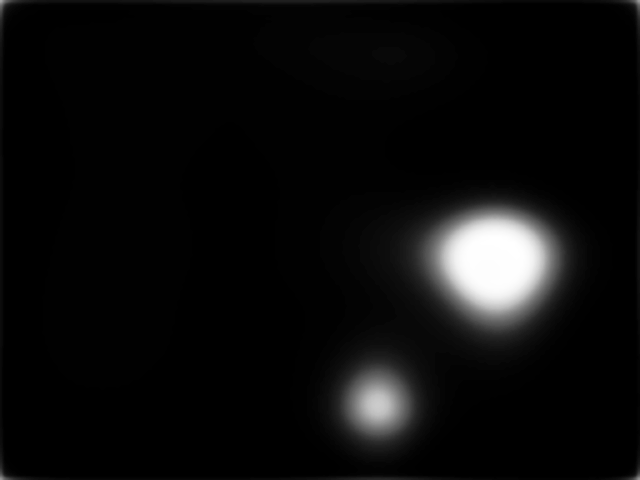}
  \includegraphics[width=2cm, height=2cm]{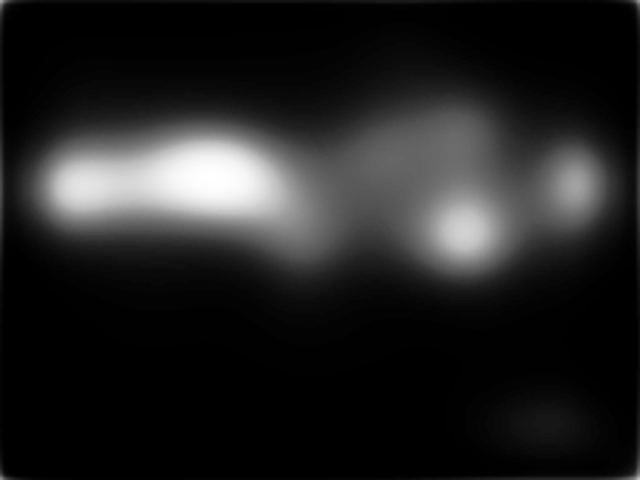}
  \includegraphics[width=2cm, height=2cm]{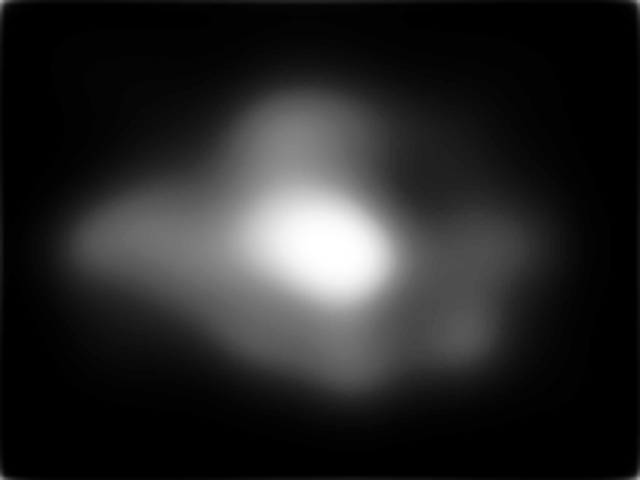}
  \includegraphics[width=2cm, height=2cm]{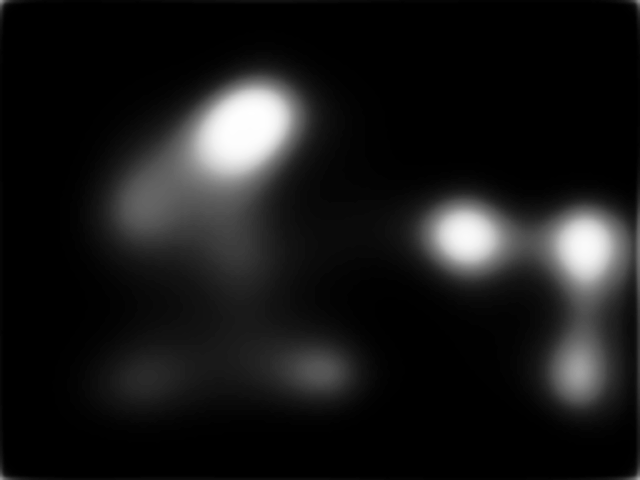}

  \caption{Comparison of SalGAN\cite{Pan_2017_SalGAN} on cleaning measures of SALICON\cite{jiang2015salicon} data attacked with DeepFool\cite{deepfool}.
             From top: Original image, ground truth, Attacked, BitDepth. JPEG, SHIELD\cite{das2018shield}, SAD
  }
  \label{fig:results_salicon}
\end{figure}

In general, DeepFool\cite{deepfool} has little to no effect on saliency prediction as is illustrated by Figure \ref{fig:results_salicon}. In this figure we see only slight difference between the original, attacked and cleaned saliency predictions. This result is further backed up by the metrics of DeepFool\cite{deepfool} across all tables. 

\begin{table*}[hbt]
\begin{tabular}{l|l|l|l|l|l}
Data: \textbf{SALICON}, Model: \textbf{BASNet}                         & EMD $\downarrow$        & CC $\uparrow$          & NSS $\uparrow$        & KLD $\downarrow$        & SIM $\uparrow$ \\
\hline
Original                        & 83.60872647 & 0.4218207896 & 0.3761202097 & 10.67353916 & 0.4060104787 \\
\hline
FGSM                           & 79.91393103 & 0.4087593555 & 0.3648214042 & 11.35947323 & 0.3891682327 \\
DeepFool                       & 83.60636636 & 0.4222988188 & 0.3763460219 & 10.67110252 & 0.40616256   \\
\hline
FGSM + Bit-depth Reduction     & \textbf{75.04985629} & 0.3049599528 & 0.2969438136 & 13.31941891 & 0.3179412484 \\
FGSM + JPEG80 Compression      & 79.73553778 & 0.4079829454 & 0.3639315665 & 11.40826893 & 0.3880238533 \\
FGSM + SHIELD                  & 79.44973525 & 0.4092005491 & 0.3637762368 & 11.43848801 & 0.3876400888 \\
FGSM + SAD (20 50 70 70 80 90)            & 79.25510666 & 0.4074067175 & 0.3620625138 & 11.51703453 & 0.3857473135 \\
FGSM + SAD (50 70 90)            & 79.89621414 & \textbf{0.4122531116} & \textbf{0.3667055368} & \textbf{11.30664825} & \textbf{0.3907471597} \\
\hline
DeepFool + Bit-depth Reduction & \textbf{78.72886619} & 0.3303083181 & 0.3210006058 & 12.43772602 & 0.3442973197 \\
DeepFool + JPEG80 Compression  & 83.44919726 & 0.421741128  & 0.3755041957 & 10.72119999 & 0.4052546024 \\
DeepFool + SHIELD              & 83.24127967 & 0.4230029285 & 0.3759891391 & 10.71790504 & 0.4055115879 \\
DeepFool + SAD (20 50 70 70 80 90)        & 83.10682231 & 0.4206542075 & 0.3742873669 & 10.78884315 & 0.403165251  \\
DeepFool + SAD (50 70 90)        & 83.54616027  & \textbf{0.4241522551} & \textbf{0.3771335781} & \textbf{10.64822292} & \textbf{0.4069490135}
\end{tabular}
  \caption{Evaluation of the BASNet\cite{BASNet} visual saliency model\cite{BASNet} on the SALICON\cite{jiang2015salicon} dataset.}
  \label{table:results_salicon_basnet}
\end{table*}

\begin{table*}[hbt]
\begin{tabular}{l|l|l|l|l|l}
Data: \textbf{ECSSD}, Model: \textbf{BASNet}                   & EMD $\downarrow$        & CC $\uparrow$          & NSS $\uparrow$        & KLD $\downarrow$        & SIM $\uparrow$ \\
\hline
Original                          & 48.07041578 & 0.9120191336 & 1.979211807 & 1.506018996 & 0.8843896985 \\
\hline
FGSM                           & 45.51845167 & 0.8434635997 & 1.829114914 & 3.207652092 & 0.8040903211 \\
DeepFool                       & 47.9678527  & 0.908826232  & 1.972466826 & 1.60269177  & 0.8807195425 \\
\hline
FGSM + Bit-depth Reduction     & \textbf{41.21447617} & 0.5987574458 & 1.297375202 & 8.744916916 & 0.5463407636 \\
FGSM + JPEG80 Compression      & 45.44040079 & 0.8403670192 & 1.823511362 & 3.267945766 & 0.8005516529 \\
FGSM + SHIELD                  & 45.26718383 & 0.8304385543 & 1.805399299 & 3.560398102 & 0.7886587977 \\
FGSM + SAD (20 50 70 70 80 90)            & 45.35846763 & 0.8506878614 & 1.85140121  & 3.170518398 & 0.8131732941 \\
FGSM + SAD (50 70 90)            & 45.93143812 & \textbf{0.8615031838} & \textbf{1.87408042}  & \textbf{2.825253487} & \textbf{0.8248550296} \\
\hline
DeepFool + Bit-depth Reduction & \textbf{43.46246487} & 0.6593744755 & 1.430215001 & 7.15671587  & 0.6113178134 \\
DeepFool + JPEG80 Compression  & 47.933426   & \textbf{0.9080747962} & \textbf{1.97224772}  & \textbf{1.61575985}  & \textbf{0.8803170323} \\
DeepFool + SHIELD              & 47.71402123 & 0.8999755979 & 1.953188896 & 1.781389356 & 0.870287478  \\
DeepFool + SAD (20 50 70 70 80 90)        & 47.33478719 & 0.9016960859 & 1.960625887 & 1.858531952 & 0.8717075586 \\
DeepFool + SAD (50 70 90)        & 47.77355919 & 0.9041004777 & 1.961497784 & 1.685516238 & 0.8761977553
\end{tabular}
  \caption{Evaluation of the BASNet\cite{BASNet} visual saliency model\cite{BASNet} on the ECSSD\cite{ECSSD} dataset.}
  \label{table:results_ecssd_basnet}
\end{table*}

\begin{table*}[hbt]
\begin{tabular}{l|l|l|l|l|l}
Data: \textbf{ECSSD}, Model: \textbf{CPD}  & EMD $\downarrow$   & CC $\uparrow$ & NSS $\uparrow$ & KLD $\downarrow$ & SIM $\uparrow$ \\
\hline
Original                          & 48.28455886 & 0.9043654799 & 1.964785576 & 1.253266454 & 0.874347806  \\
\hline
FGSM                           & 45.36623833 & 0.8253191113 & 1.793251872 & 3.001737833 & 0.7846102118 \\
DeepFool                       & 48.16762985 & 0.9003679752 & 1.957540512 & 1.323252082 & 0.8697779179 \\
\hline
FGSM + Bit-depth Reduction     & \textbf{37.93778128} & 0.5365927815 & 1.133271813 & 9.139689445 & 0.4919550121 \\
FGSM + JPEG80 Compression      & 45.04148904 & 0.8208998442 & 1.786798239 & 3.14307785  & 0.7796351314 \\
FGSM + SHIELD                  & 44.63645556 & 0.8076060414 & 1.756557226 & 3.521080494 & 0.7643808722 \\
FGSM + SAD (20 50 70 70 80 90)            & 44.82445048 & 0.816835165  & 1.78500545  & 3.504522562 & 0.7739418745 \\
FGSM + SAD (50 70 90)            & 45.2943999  & \textbf{0.8305669427} & \textbf{1.812999964} & \textbf{3.0791049} & \textbf{0.7902354002} \\
\hline
DeepFool + Bit-depth Reduction & \textbf{40.7708216}  & 0.62786448   & 1.342035532 & 6.702296734 & 0.5820772648 \\
DeepFool + JPEG80 Compression  & 47.96485411 & \textbf{0.8997527361} & \textbf{1.956662774} & \textbf{1.369203806} & \textbf{0.869066}     \\
DeepFool + SHIELD              & 47.49587288 & 0.8849500418 & 1.92660892  & 1.676331162 & 0.8517687917 \\
DeepFool + SAD (20 50 70 70 80 90)        & 47.21730354 & 0.8871904016 & 1.92954421  & 1.74879241  & 0.8548846841 \\
DeepFool + SAD (50 70 90)       & 47.65713406 & 0.892139554  & 1.939523816 & 1.546016574 & 0.8613493443
\end{tabular}
  \caption{Evaluation of the CPD\cite{wu2019cascaded} saliency visual saliency model\cite{wu2019cascaded} on the ECSSD\cite{ECSSD} dataset.}
  \label{table:results_ecssd_cpd}
\end{table*}

\begin{table*}[hbt]
\begin{tabular}{l|l|l|l|l|l}
Data: \textbf{SALICON}, Model: \textbf{CPD}                        & EMD $\downarrow$        & CC $\uparrow$          & NSS $\uparrow$        & KLD $\downarrow$        & SIM $\uparrow$\\
\hline
Salicon                        & 84.03098486 & 0.464445889  & 0.4046932757 & 9.461788177 & 0.4308495224 \\
FGSM                           & 77.17695558 & 0.4409204125 & 0.3802604973 & 10.84904957 & 0.3955992162 \\
DeepFool                       & 83.99931418 & 0.4644609392 & 0.4047108293 & 9.465325356 & 0.4308281243 \\
\hline
FGSM + Bit-depth Reduction     & \textbf{71.81423414} & 0.3703564703 & 0.344774574  & 11.47888279 & 0.3569065928 \\
FGSM + JPEG80 Compression      & 76.72632067 & \textbf{0.43930161}   & \textbf{0.378226012}  & \textbf{10.95398521} & \textbf{0.3932281733} \\
FGSM + SHIELD                  & 76.17508536 & 0.4379900992 & 0.3765870929 & 11.05368805 & 0.3907161355 \\
FGSM + SAD (20 50 70 70 80 90)            & 76.01439047 & 0.4356130958 & 0.3746709526 & 11.19257164 & 0.3878324628 \\
FGSM + SAD (50 70 90)             & 76.42406584 & 0.438691169  & 0.3776216805 & 11.02315617 & 0.3915502429 \\
\hline
DeepFool + Bit-depth Reduction & \textbf{76.4548955}3 & 0.416684866  & 0.3889657855 & 10.1579113  & 0.3980270326 \\
DeepFool + JPEG80 Compression  & 83.7110553  & 0.4636833072 & \textbf{0.4037819207} & \textbf{9.535971642} & \textbf{0.4293003976} \\
DeepFool + SHIELD              & 82.80526771 & \textbf{0.4643773437} & 0.4027466178 & 9.67798233  & 0.4266972542 \\
DeepFool + SAD (20 50 70 70 80 90)        & 82.65074847 & 0.4604454637 & 0.3989365101 & 9.831030846 & 0.4227913916 \\
DeepFool + SAD (50 70 90)         & 83.29619378 & 0.4635473192 & 0.4021643996 & 9.622299194 & 0.4274015129
\end{tabular}
  \caption{Evaluation of the CPD \cite{wu2019cascaded} visual saliency model\cite{wu2019cascaded} on the SALICON\cite{jiang2015salicon} dataset.}
  \label{table:results_salicon_cpd}
\end{table*}

\section{Conclusion and Future Work}

With adversarial attacks increasing in popularity and constantly evolving, new defenses are continuously being counteracted by new methods of attack.
In this work, we presented a new method for defense against adversarial images which is based upon visual saliency estimation.
In comparison with existing localized and global approaches, our method is a strategically applied defense.
Our targeted approach demonstrates better reduction of adversarial distortions while preserving salient content of the original data.
Our proposed SAD model outperforms existing countermeasures in a range of standard saliency metrics.

While SAD has been proven effective, there are still many different areas to explore.
In future work, we will look to optimize saliency thresholds as well as the back-end saliency model, to further improve the results of SAD. 
Further analysis of the effectiveness of our model will be explored in comparison with a growing number of state-of-the-art defenses on additional saliency datasets, and can be analyzed in terms of the classification of images across similar phases - before attack, during attack, and after cleaned.

\clearpage
{\small
\bibliographystyle{ieee}
\bibliography{egbib}

\begin{thebibliography}{10}\itemsep=-1pt

\bibitem{athalye2018synthesizing}
A.~Athalye, L.~Engstrom, A.~Ilyas, and K.~Kwok.
\newblock Synthesizing robust adversarial examples.
\newblock In {\em International Conference on Machine Learning}, pages
  284--293, 2018.

\bibitem{mit-saliency-benchmark}
Z.~Bylinskii, T.~Judd, F.~Durand, A.~Oliva, and A.~Torralba.
\newblock Mit saliency benchmark.
\newblock http://saliency.mit.edu/.

\bibitem{CWL2}
N.~{Carlini} and D.~{Wagner}.
\newblock Towards evaluating the robustness of neural networks.
\newblock In {\em 2017 IEEE Symposium on Security and Privacy (SP)}, pages
  39--57, May 2017.

\bibitem{das2018shield}
N.~Das, M.~Shanbhogue, S.-T. Chen, F.~Hohman, S.~Li, L.~Chen, M.~E. Kounavis,
  and D.~H. Chau.
\newblock Shield: Fast, practical defense and vaccination for deep learning
  using jpeg compression.
\newblock In {\em Proceedings of the 24th ACM SIGKDD International Conference
  on Knowledge Discovery \& Data Mining}, pages 196--204. ACM, 2018.

\bibitem{imagenet_cvpr09}
J.~Deng, W.~Dong, R.~Socher, L.-J. Li, K.~Li, and L.~Fei-Fei.
\newblock {ImageNet: A Large-Scale Hierarchical Image Database}.
\newblock In {\em CVPR09}, 2009.

\bibitem{quilting}
A.~A. Efros and W.~T. Freeman.
\newblock Image quilting for texture synthesis and transfer.
\newblock In {\em Proceedings of the 28th annual conference on Computer
  graphics and interactive techniques}, pages 341--346. ACM, 2001.

\bibitem{adv_sal2019}
A.~Fernandez.
\newblock On the salience of adversarial examples.
\newblock In {\em 14th International Symposium on Visual Computing (ISVC)},
  2019.

\bibitem{FGSM}
I.~Goodfellow, J.~Shlens, and C.~Szegedy.
\newblock Explaining and harnessing adversarial examples.
\newblock In {\em International Conference on Learning Representations}, 2015.

\bibitem{grosse2017adversarial}
K.~Grosse, N.~Papernot, P.~Manoharan, M.~Backes, and P.~McDaniel.
\newblock Adversarial examples for malware detection.
\newblock In {\em European Symposium on Research in Computer Security}, pages
  62--79. Springer, 2017.

\bibitem{jiang2015salicon}
M.~Jiang, S.~Huang, J.~Duan, and Q.~Zhao.
\newblock Salicon: Saliency in context.
\newblock In {\em The IEEE Conference on Computer Vision and Pattern
  Recognition (CVPR)}, June 2015.

\bibitem{IFGSM}
A.~Kurakin, I.~Goodfellow, and S.~Bengio.
\newblock Adversarial examples in the physical world.
\newblock {\em arXiv preprint arXiv:1607.02533}, 2016.

\bibitem{kurakin2016adversarial}
A.~Kurakin, I.~Goodfellow, and S.~Bengio.
\newblock Adversarial machine learning at scale.
\newblock {\em arXiv preprint arXiv:1611.01236}, 2016.

\bibitem{liu2018picanet}
N.~Liu, J.~Han, and M.-H. Yang.
\newblock Picanet: Learning pixel-wise contextual attention for saliency
  detection.
\newblock In {\em Proceedings of the IEEE Conference on Computer Vision and
  Pattern Recognition}, pages 3089--3098, 2018.

\bibitem{meng2017magnet}
D.~Meng and H.~Chen.
\newblock Magnet: a two-pronged defense against adversarial examples.
\newblock In {\em Proceedings of the 2017 ACM SIGSAC Conference on Computer and
  Communications Security}, pages 135--147. ACM, 2017.

\bibitem{deepfool}
S.-M. Moosavi-Dezfooli, A.~Fawzi, and P.~Frossard.
\newblock Deepfool: a simple and accurate method to fool deep neural networks.
\newblock In {\em Proceedings of the IEEE conference on computer vision and
  pattern recognition}, pages 2574--2582, 2016.

\bibitem{Pan_2017_SalGAN}
J.~Pan, E.~Sayrol, X.~G.-i. Nieto, C.~C. Ferrer, J.~Torres, K.~McGuinness, and
  N.~E. OConnor.
\newblock Salgan: Visual saliency prediction with adversarial networks.
\newblock In {\em CVPR Scene Understanding Workshop (SUNw)}, 2017.

\bibitem{Papernot2015DistillationAA}
N.~Papernot, P.~D. McDaniel, X.~Wu, S.~Jha, and A.~Swami.
\newblock Distillation as a defense to adversarial perturbations against deep
  neural networks.
\newblock {\em 2016 IEEE Symposium on Security and Privacy (SP)}, pages
  582--597, 2015.

\bibitem{BASNet}
X.~Qin, Z.~Zhang, C.~Huang, C.~Gao, M.~Dehghan, and M.~Jagersand.
\newblock Basnet: Boundary-aware salient object detection.
\newblock In {\em The IEEE Conference on Computer Vision and Pattern
  Recognition (CVPR)}, June 2019.

\bibitem{qin2019imperceptible}
Y.~Qin, N.~Carlini, G.~Cottrell, I.~Goodfellow, and C.~Raffel.
\newblock Imperceptible, robust, and targeted adversarial examples for
  automatic speech recognition.
\newblock In {\em International Conference on Machine Learning}, pages
  5231--5240, 2019.

\bibitem{sharif2016accessorize}
M.~Sharif, S.~Bhagavatula, L.~Bauer, and M.~K. Reiter.
\newblock Accessorize to a crime: Real and stealthy attacks on state-of-the-art
  face recognition.
\newblock In {\em Proceedings of the 2016 ACM SIGSAC Conference on Computer and
  Communications Security}, pages 1528--1540. ACM, 2016.

\bibitem{ECSSD}
J.~{Shi}, Q.~{Yan}, L.~{Xu}, and J.~{Jia}.
\newblock Hierarchical image saliency detection on extended cssd.
\newblock {\em IEEE Transactions on Pattern Analysis and Machine Intelligence},
  38(4):717--729, April 2016.

\bibitem{vgg16}
K.~Simonyan and A.~Zisserman.
\newblock Very deep convolutional networks for large-scale image recognition.
\newblock {\em arXiv preprint arXiv:1409.1556}, 2014.

\bibitem{Wang_2017_CVPR}
L.~Wang, H.~Lu, Y.~Wang, M.~Feng, D.~Wang, B.~Yin, and X.~Ruan.
\newblock Learning to detect salient objects with image-level supervision.
\newblock In {\em The IEEE Conference on Computer Vision and Pattern
  Recognition (CVPR)}, July 2017.

\bibitem{wu2019cascaded}
Z.~Wu, L.~Su, and Q.~Huang.
\newblock Cascaded partial decoder for fast and accurate salient object
  detection.
\newblock In {\em Proceedings of the IEEE Conference on Computer Vision and
  Pattern Recognition}, pages 3907--3916, 2019.

\end{thebibliography}
}
\end{document}